\documentclass{article}

\usepackage[main, final]{neurips_2026}
\usepackage{amssymb}
\usepackage{amsmath}
\usepackage{booktabs}
\usepackage{multirow}
\usepackage[table]{xcolor}

\usepackage[utf8]{inputenc}
\usepackage{booktabs}      
\usepackage{graphicx}      

\definecolor{groupblue}{HTML}{F4F7FB} 
\definecolor{oursblue}{HTML}{E2EDFA}  

\usepackage[utf8]{inputenc} 
\usepackage[T1]{fontenc}    
\usepackage{hyperref}       
\usepackage{url}            
\usepackage{booktabs}       
\usepackage{amsfonts}       
\usepackage{nicefrac}       
\usepackage{microtype}      
\usepackage{xcolor}         

\title{SceneScaffold: Active Scene-State Construction\\for Unified 3D Scene Understanding}

\author{%
  Xiangqi Li$^{1,2}$\quad
  Libo Huang$^{1}$\thanks{Corresponding authors.}\quad
  Jiarui Zhao$^{2}$\quad
  Weilun Feng$^{1,2}$ \\
  \bf Chuanguang Yang$^{1}$\qquad
  Zhulin An$^{1}$\footnotemark[1]\qquad
  Yongjun Xu$^{1}$ \\[4pt]
  $^{1}$State Key Laboratory of AI Safety, Institute of Computing Technology, Chinese Academy of Sciences \\
  $^{2}$University of Chinese Academy of Sciences \\
  \texttt{\{lixiangqi24s,huanglibo,fengweilun24s,yangchuanguang,anzhulin,xyj\}@ict.ac.cn} \\
  \texttt{zhaojiarui231@mails.ucas.ac.cn} \\
  Code: \url{https://github.com/lixiangqi707/SceneScaffold}
}

\begin{document}

\maketitle

\begin{abstract}

Recent 3D large multimodal models (3D-LMMs) rely on a visual bottleneck to compress complex 3D scene evidence into a limited number of visual tokens compatible with large language models (LLMs).
Current visual bottlenecks, however, often passively compress heterogeneous 3D evidence into a homogeneous object-centric token sequence, leaving the spatial organization of the scene under-represented.
This under-representation forces the LLM to recover spatial relations from a flattened token sequence, leading to unstable reasoning in relation-intensive and spatially ambiguous scenes.
To address this issue, we propose \textbf{SceneScaffold}, an active scene-state construction framework for unified 3D scene understanding.
SceneScaffold reformulates the visual bottleneck from a passive feature compressor into an active scene organizer, constructing a role-aware spatial scaffold before language reasoning.
Specifically, SceneScaffold organizes superpoint-level visual evidence into scene-state components with distinct structural roles: entity states preserve core object semantics, scene-frame states maintain spatial references via boundary and region anchors, relation states encode object-environment interaction cues, and a global summary provides compact context.
Through this role-aware construction, SceneScaffold provides the LLM with a spatially organized scene representation before language reasoning.
Experiments on unified 3D scene understanding tasks, including 3D visual grounding, question answering, and dense captioning, demonstrate the effectiveness of SceneScaffold, while diagnostic results further show its applicability to relation-intensive and spatially ambiguous cases.

\end{abstract}

\section{Introduction}

\begin{figure*}[t]
    \centering
    \includegraphics[width=\textwidth]{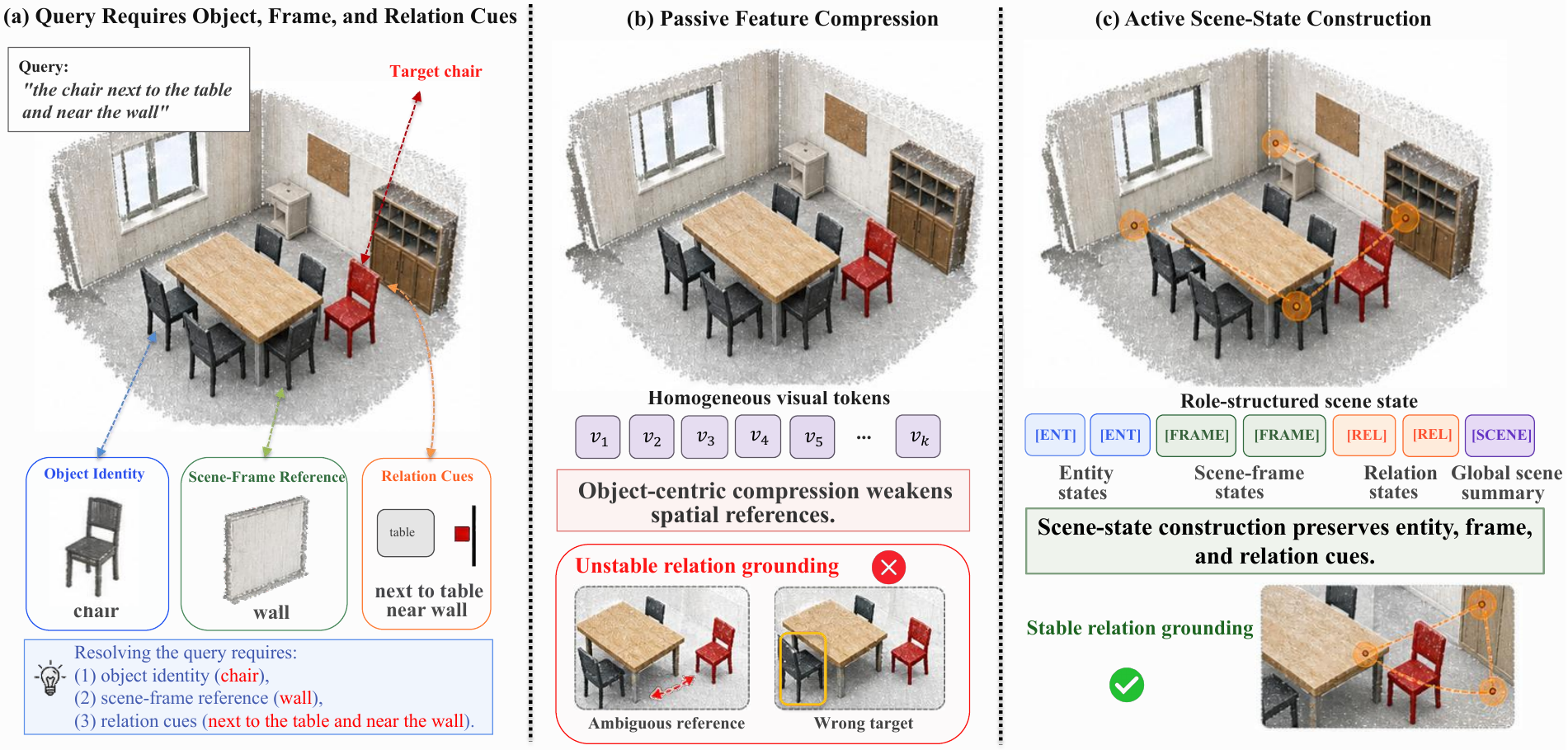}
    \caption{
    \textbf{Motivation of SceneScaffold.}
    Relation-intensive 3D queries require object identity, scene-frame references, and relation cues. 
    Under a fixed visual bottleneck, passive feature compression tends to preserve object-centric fragments while weakening spatial references, which can lead to unstable relation grounding. 
    In contrast, SceneScaffold organizes visual evidence into a role-structured scene state before language reasoning, explicitly preserving entity evidence, scene-frame references, relation cues, and global scene context.
    }
    \label{fig:motivation}
    \vspace{-0.5em}
\end{figure*}

Large Language Models (LLMs) have emerged as the central interface for general semantic reasoning and human-agent interaction~\cite{brown2020language,openai2023gpt4,an2024multimodality,jiang2025unlocking,huang2025r,feng2026mpqdmv2}. Extending this formidable capability to 3D physical environments is a critical step toward embodied intelligence and real-world spatial interaction~\cite{an2025embodied,huang2023leo,kawaharazuka2024real,li2025geometric}. Consequently, recent 3D Large Multimodal Models (3D-LMMs) connect 3D scene representations with natural language within a unified LLM-based architecture~\cite{liu2024uni3d,hong2023_3dllm,chen2024_ll3da}. These models are designed to handle diverse scene-level tasks simultaneously, including 3D visual grounding~\cite{achlioptas2020referit3d,zhang2023multi3drefer}, question answering~\cite{azuma2022scanqa,ma2023sqa3d}, and dense captioning~\cite{chen2021scan2cap,chen2023vote2capdetr}.
Since raw 3D scenes contain a massive number of points and candidate regions, a visual bottleneck is indispensable to compress complex 3D representations into a strictly limited number of visual tokens for the LLM~\cite{deng2025_3dllava,tang2025scenes}.

This bottleneck is especially consequential for 3D scenes, which are continuous physical environments characterized by metric scale, spatial extents, boundaries, and object-environment interactions~\cite{dai2017scannet,chang2017matterport3d,chen2024_grounded3dllm}. 
As illustrated in Figure~\ref{fig:motivation}, resolving a relation-intensive 3D query requires object identity to be grounded together with scene-frame references and relation cues.
These spatial references define object locations and their relations to the surroundings, and are essential for reliable 3D understanding~\cite{azuma2022scanqa,ma2023sqa3d,zhou2023uni3d,qi2024_gpt4point,chen2020scanrefer,zhao2021_3dvg,li2026pointuq}.
However, current 3D-LMMs typically instantiate the visual bottleneck as a homogeneous feature compressor~\cite{chen2024_ll3da,deng2025_3dllava,tang2025scenes}. 
Large-scale 3D inputs are selected or pooled into a flat token sequence biased toward semantically salient object-centric evidence~\cite{deng2025_3dllava,tang2025scenes}.
Consequently, critical spatial references, such as scene boundaries, occupied regions, and coarse layout cues, can be under-represented in the limited interface~\cite{zhao2021_3dvg,luo2022_3dsps,wu2023eda}.
The object-centric compression process degrades the coherent scene state into a set of isolated object fragments.
It also disrupts the collaborative division of labor between the visual and language modalities. 
When the visual side provides only weakly structured fragments, the language model has to recover scene references, object positions, and spatial relations from a flattened token sequence.
This shifted responsibility creates a modal role mismatch: the visual side mainly filters visual evidence, while the language side is forced to perform implicit compensatory reasoning over a one-dimensional sequence to reconstruct missing 3D relations.
Such a modal role mismatch is particularly detrimental to unified 3D scene understanding, where multi-task reasoning shares the same physical environment but demands multi-granularity spatial states. 
For instance, question answering requires global layout comprehension~\cite{azuma2022scanqa,ma2023sqa3d}, whereas visual grounding requires fine-grained spatial disambiguation among densely distributed, visually similar entities~\cite{chen2020scanrefer,achlioptas2020referit3d,zhang2023multi3drefer,zhao2021_3dvg}. 
Forcing the LLM to simultaneously shoulder the dual burden of semantic reasoning and compensatory spatial-structure recovery tends to cause relational confusion, reference drift, and unstable localization as scene complexity increases.

To address these limitations, we argue for a paradigm shift: the visual bottleneck should evolve from a passive feature compressor into an \textit{active scene organizer}. It should explicitly clarify the structural roles of different types of visual evidence and assemble a coarse spatial-reference scaffold before the features enter the LLM. Based on this view, we propose \textbf{SceneScaffold}, an active scene-state construction framework for unified 3D scene understanding.
Under a fixed cross-modal token budget, SceneScaffold organizes superpoint-level visual evidence into four role-grounded components: \textit{entity states} preserve core object semantics and local details; \textit{scene-frame states} maintain spatial references via boundary and region anchors; \textit{relation states} explicitly encode object-object and object-environment interaction cues; and a \textit{global scene summary} distills compact context for multi-task reuse. 
Through this role-aware construction and budget-preserving serialization mechanism,
the downstream language model receives a shared scene representation whose basic
spatial organization has already been formed on the visual side.

Our main contributions are summarized as follows. 
First, we analyze the visual bottleneck of unified 3D-LMMs and reveal a modal role mismatch: object-centric compression under-represents spatial references and shifts implicit compensatory spatial reasoning to the language side. 
Second, we reformulate the visual bottleneck from passive feature compression to active scene-state construction, organizing entity evidence, scene-frame references, and relation cues into a role-structured scene-state interface under a fixed cross-modal token budget. 
Third, we design \textbf{SceneScaffold}, a unified framework that constructs entity states, scene-frame states, relation states, and a global scene summary as a shared scene representation for multiple 3D tasks. 
Finally, we evaluate SceneScaffold on 3D visual grounding, question answering, and dense captioning, demonstrating its effectiveness for relation-intensive and spatially ambiguous 3D scene understanding.

\section{Related Work}

\subsection{Unified 3D Scene Understanding and 3D LMMs}

3D vision-language understanding has evolved from task-specific modeling toward unified scene-level understanding~\cite{tang2025scenes,deng2025_3dllava,hong2023_3dllm,fu2024_scenellm}. Early works study individual tasks, including 3D visual grounding in ScanRefer~\cite{chen2020scanrefer} and ReferIt3D~\cite{achlioptas2020referit3d}, dense captioning in Scan2Cap~\cite{chen2021scan2cap}, and question answering in ScanQA~\cite{azuma2022scanqa} and SQA3D~\cite{ma2023sqa3d}. Unified models such as 3DJCG~\cite{cai2022_3djcg} and UniT3D~\cite{chen2023unit3d} jointly handle grounding and captioning, but still rely heavily on task-specific heads.
Recent 3D large multimodal models connect 3D scene representations to LLMs for open-ended interaction. 3D-LLM~\cite{hong2023_3dllm} injects lifted 3D features into LLMs, LL3DA~\cite{chen2024_ll3da} introduces visual interactive instruction tuning, PointLLM~\cite{xu2024_pointllm} aligns point clouds with LLMs, Chat-Scene~\cite{huang2024_chatscene} uses object identifiers for scene-level interaction, and Scene-LLM~\cite{fu2024_scenellm} extends language models to 3D reasoning. More recent methods further improve the 3D-LMM interface: 3D-LLaVA~\cite{deng2025_3dllava} integrates visual token selection, prompt encoding, and mask decoding with an Omni Superpoint Transformer; PerLA~\cite{mei2025_perla} enhances local-global visual representations; LSceneLLM~\cite{zhi2025lscenellm} adaptively selects task-relevant regions; Scenes as Tokens~\cite{tang2025scenes} constructs holistic scene tokens with a multi-scale NDT tokenizer.
Although these models advance unified 3D scene understanding, most still formulate the cross-modal interface as a finite set of tokens, proposals, or selected features~\cite{an2026parameterized}. Their interfaces are typically designed around feature selection, alignment, or tokenization, while the internal role structure of compressed 3D evidence is less explicitly studied. In contrast, we revisit the visual bottleneck itself and argue that unified 3D scene understanding requires a role-preserving shared scene state rather than a homogeneous collection of visual fragments.

\subsection{Visual Bottlenecks and Scene Representation}

Visual bottlenecks are widely used to reduce the dimensionality of visual inputs before language reasoning~\cite{deng2025_3dllava,lin2025multi,zhao2026rsiat}.
In video LLMs, DynFocus encodes question-relevant frames in greater detail under a limited token budget~\cite{han2025dynfocus}.
DyCPO distinguishes visual exploration from answer-relevant reasoning tokens during optimization~\cite{han2026dycpo}.
Visual-enhanced depth scaling allocates additional reasoning steps to complex tokens~\cite{han2026vedas}.
In 2D multimodal models, LLaVA~\cite{liu2023llava} uses projection layers to align visual features with LLMs, while BLIP-2~\cite{li2023blip2} adopts query-based connectors.
Multimodal Wasserstein barycenter alignment provides a shared representation space across multiple modalities~\cite{tang2026barybind}.
In all-in-one image restoration, barycenter and residual spaces separate degradation-invariant content from degradation-specific information~\cite{tang2026learning}.
Efficient vision transformers further reduce token redundancy through learned token selection in TokenLearner~\cite{ryoo2021tokenlearner}, dynamic token sparsification in DynamicViT~\cite{rao2021dynamicvit}, token reorganization in EViT~\cite{liang2022evit}, and token merging in ToMe~\cite{bolya2023token}.
Image-restoration transformers also use adaptive sparse attention to suppress irrelevant interactions~\cite{Zhou2025LearningAA}.
Hierarchical attention diversifies feature learning and reduces head redundancy~\cite{Zhou2025DevilII}, while high-frequency injection helps retain local detail~\cite{Zhou2026RethinkingTI}.
However, 3D scene evidence is inherently heterogeneous: object surfaces, scene boundaries, layout regions, object-environment references, and object-object relations play different roles. Spatially grounded 3D grounding methods, such as 3DVG-Transformer~\cite{zhao2021_3dvg}, BUTD-DETR~\cite{jain2022butdetr}, 3D-SPS~\cite{luo2022_3dsps}, and EDA~\cite{wu2023eda}, model spatial context for grounding, but their relation modeling is usually tied to specific task objectives rather than serving as a reusable interface for LLM-based multi-task understanding.

Our work differs from prior token compression and task-specific relation modeling. We do not simply select more salient object tokens under a fixed budget. Instead, we formulate the 3D visual bottleneck as role-preserving scene-state construction, where object-core evidence, scene-frame anchors, and relation-bearing cues are explicitly organized before language reasoning. This provides LLMs with a reusable 3D scene state for visual grounding, question answering, and dense captioning.

\newcommand{\method}{SceneScaffold}

\section{Method}
\label{sec:method}

To address the modal role mismatch caused by homogeneous object-centric compression, we enable the visual side to organize entity evidence, scene-frame references, and relation cues into a compact scene state before language reasoning.
We propose \method{}, a role-constrained scene-state construction framework under a fixed visual-token budget.
Given visual evidence from a 3D scene, \method{} organizes the visual bottleneck into a multi-task shared scene-state interface. The resulting visual input exposes object evidence, spatial references, and relational cues to the LLM.

\begin{figure*}[t]
    \centering
    \includegraphics[width=\textwidth]{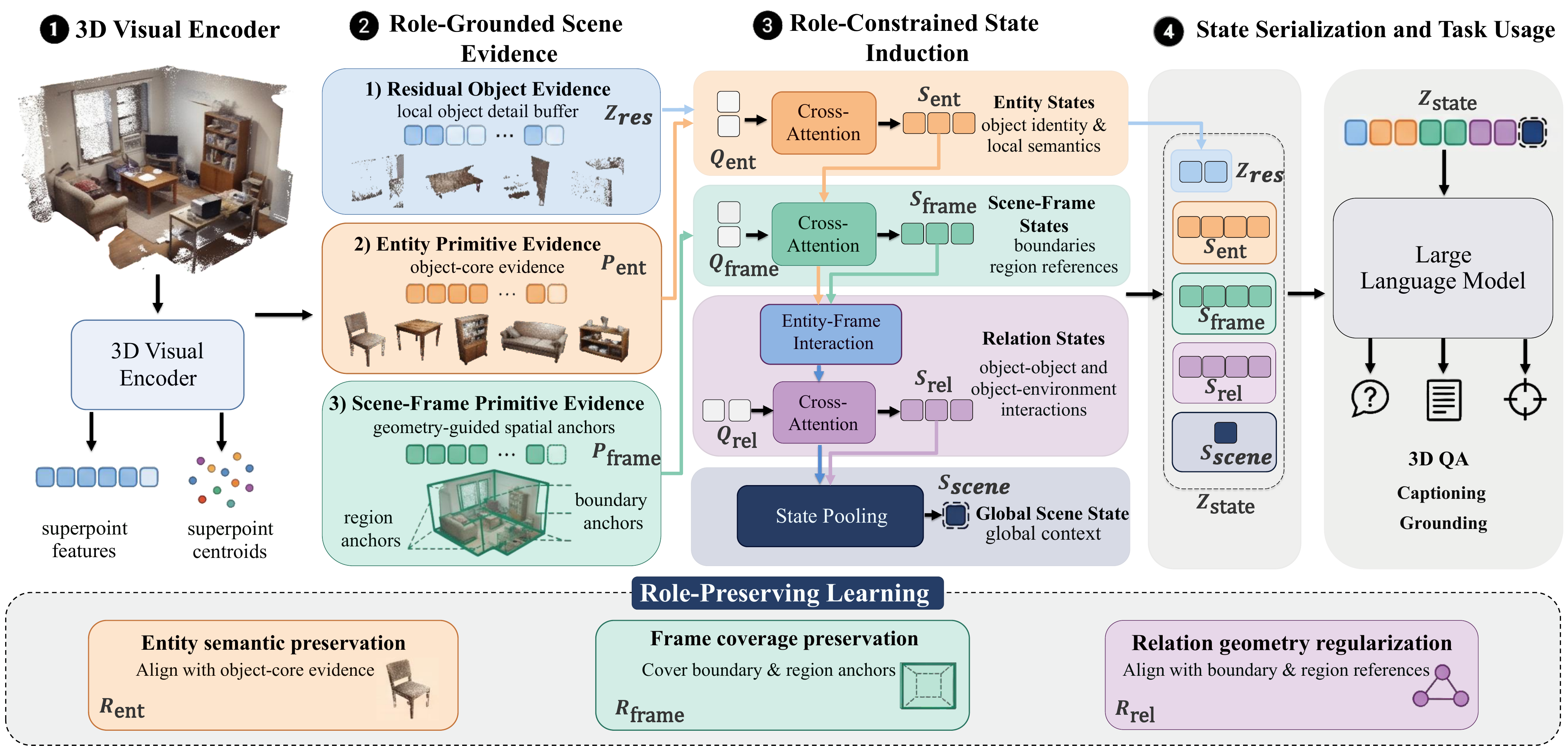}
    \caption{
    \textbf{Overview of SceneScaffold.}
    Given a 3D scene, the visual encoder extracts superpoint features and centroids. SceneScaffold first organizes them into a residual object buffer, entity primitive evidence, and geometry-guided scene-frame evidence. It then induces entity states, scene-frame states, relation states, and a global scene summary through role-constrained state readers, and serializes the resulting states as a fixed-budget visual interface for the LLM. Role-preserving learning regularizes entity, frame, and relation states with object semantics, spatial coverage, and coarse relation geometry, respectively.
    }
    \label{fig:method_overview}
\end{figure*}

As illustrated in Figure~\ref{fig:method_overview}, the forward pipeline of \method{} consists of four stages: 3D visual encoding, role-grounded scene evidence construction, role-constrained state induction, and state serialization for task usage. Role-preserving learning is applied during training to maintain the structural responsibilities of the induced states.

\subsection{Scene-State Interface under a Fixed Visual Budget}
\label{sec:state_interface}

Given an input point cloud $X$, a 3D visual encoder first extracts superpoint-level features and their physical centroids:
\begin{equation}
(H,U)=\mathcal{E}_{\mathrm{3D}}(X),
\end{equation}
where $H=\{h_j\}_{j=1}^{M}$ denotes $M$ superpoint features, and $U=\{u_j\}_{j=1}^{M}$ denotes their corresponding 3D centroids. Unified 3D scene understanding requires this dense superpoint-level evidence to be transformed into a length-limited visual input for the LLM. 
\method{} formulates this limited visual interface as a fixed-length sequence with $K$ visual tokens:
\begin{equation}
Z_{\mathrm{state}}
=
\Pi
\left(
Z_{\mathrm{res}},
S_{\mathrm{ent}},
S_{\mathrm{frame}},
S_{\mathrm{rel}},
S_{\mathrm{scene}}
\right),
\quad
|Z_{\mathrm{state}}|=K .
\end{equation}
Here, $Z_{\mathrm{res}}$ is a residual object buffer that preserves fine-grained object details from the original visual connector. $S_{\mathrm{ent}}$ denotes entity states that encode core object identity and semantics. $S_{\mathrm{frame}}$ denotes scene-frame states that maintain spatial references such as boundaries and regions. $S_{\mathrm{rel}}$ denotes relation states that model spatial interactions between object evidence and scene-frame references. $S_{\mathrm{scene}}$ denotes a global scene summary, and $\Pi(\cdot)$ is the final serialization function.

The fixed budget serves as a constraint on the cross-modal interface. Under this constraint, \method{} grounds different components of the visual interface in distinct evidence sources and induction pathways: residual tokens preserve local object details, entity states read object-core evidence, frame states read geometric spatial anchors, and relation states are induced from entity-frame interactions. This design organizes the limited interface into a role-structured scene-state representation before it is consumed by the LLM.

\subsection{Role-Grounded Scene Evidence}
\label{sec:role_grounded_evidence}

\method{} first constructs scene evidence from different sources. The role of each state is grounded at the evidence level: entity states read object-core evidence, scene-frame states read geometric spatial evidence, and the residual object buffer preserves fine-grained object information from the original visual connector.

\paragraph{Entity evidence.}
For entity states, we use a lightweight entity scoring head to extract object-core evidence from the global superpoint features. Let $f_{\mathrm{ent}}$ be the entity scoring function. The entity evidence is defined as
\begin{equation}
\mathcal{P}_{\mathrm{ent}}
=
\mathrm{TopK}
\left(
H,
f_{\mathrm{ent}}(H),
K_{\mathrm{ent}}
\right).
\end{equation}
In our implementation, $f_{\mathrm{ent}}$ is a lightweight shared MLP that maps each superpoint feature to a scalar entity score, without language conditioning.
Here, $K_{\mathrm{ent}}$ denotes the number of selected entity evidence tokens.
$\mathcal{P}_{\mathrm{ent}}$ gathers representative evidence for object identity and local semantics in the current scene. It serves as the evidence source used by the entity reader to induce a fixed number of entity states.

\paragraph{Scene-frame evidence.}
For scene-frame states, we construct geometric spatial evidence from the superpoint centroids $U$. Specifically, we generate two types of spatial anchors from the horizontal scene extent. The first type consists of anchors pooled from the left, right, front, and back boundary bands. The second type consists of region anchors aggregated from occupied spatial grid cells. Let $\mathcal{A}_{\mathrm{bd}}$ and $\mathcal{A}_{\mathrm{reg}}$ denote the boundary anchors and region anchors, respectively. The scene-frame evidence is written as
\begin{equation}
\mathcal{P}_{\mathrm{frame}}
=
\mathcal{A}_{\mathrm{bd}}
\cup
\mathcal{A}_{\mathrm{reg}} .
\end{equation}
Since this evidence source is determined by geometric locations, it enables scene-frame states to cover spatial references such as boundaries and global occupancy patterns.

In parallel, \method{} keeps a subset of the original visual connector outputs as $Z_{\mathrm{res}}$ in the final visual sequence. This residual object buffer preserves local details and helps retain discriminative appearance cues, especially for small objects. Therefore, \method{} forms three complementary inputs at the evidence level: residual object evidence for detail preservation, object-core evidence for entity induction, and scene-frame evidence for spatial organization.

\subsection{Role-Constrained State Induction}
\label{sec:state_induction}

\method{} applies learnable state readers on top of role-grounded evidence, so that each state is conditioned by the current scene, its evidence source, and its induction path.
Given an evidence set $\mathcal{P}$ and a learnable role reader $Q$, the state induction operator is abstracted as
\begin{equation}
S=\Phi(Q,\mathcal{P}),
\end{equation}
where $\Phi(\cdot)$ is implemented with a lightweight cross-attention reader, followed by a residual connection and layer normalization. We provide the concrete reader structure in Appendix~\ref{app:state_reader_serialization}. The learnable reader $Q$ acts as a role-specific evidence reader, while the output state $S$ is conditioned on the current scene evidence.

Entity states and scene-frame states are induced from their corresponding evidence sources:
\begin{equation}
S_{\mathrm{ent}}
=
\Phi_{\mathrm{ent}}
\left(
Q_{\mathrm{ent}},
\mathcal{P}_{\mathrm{ent}}
\right),
\end{equation}
\begin{equation}
S_{\mathrm{frame}}
=
\Phi_{\mathrm{frame}}
\left(
Q_{\mathrm{frame}},
\mathcal{P}_{\mathrm{frame}}
\right).
\end{equation}
The former aggregates core object information, while the latter aggregates spatial boundary and region references.
Relation states are induced after entity states and scene-frame states have been formed:
\begin{equation}
S_{\mathrm{rel}}
=
\Phi_{\mathrm{rel}}
\left(
Q_{\mathrm{rel}},
[
S_{\mathrm{ent}};
S_{\mathrm{frame}}
]
\right).
\end{equation}
This induction pathway targets the compositional nature of 3D scene relations, where object identity is coupled with scene references such as boundary proximity, inter-object adjacency, and regional localization.
By reading $[S_{\mathrm{ent}};S_{\mathrm{frame}}]$, $S_{\mathrm{rel}}$ aggregates interaction cues among objects and between objects and the environment.

Finally, the global scene summary is obtained by aggregating the constructed role-specific states:
\begin{equation}
S_{\mathrm{scene}}
=
\phi_{\mathrm{scene}}
\left(
[
\mathrm{Pool}(S_{\mathrm{ent}});
\mathrm{Pool}(S_{\mathrm{frame}});
\mathrm{Pool}(S_{\mathrm{rel}})
]
\right).
\end{equation}
This summary provides compact global context for downstream reasoning. In this way, \method{} constructs role-structured scene states from superpoint-level visual evidence.

\subsection{State Serialization and Task Usage}
\label{sec:state_serialization}

After state construction, \method{} projects the residual object buffer and all role-specific states into the LLM embedding space and serializes them as a unified visual interface:
\begin{equation}
Z_{\mathrm{state}}
=
\left[
\psi(Z_{\mathrm{res}})+r_{\mathrm{res}};
\psi(S_{\mathrm{ent}})+r_{\mathrm{ent}};
\psi(S_{\mathrm{frame}})+r_{\mathrm{frame}};
\psi(S_{\mathrm{rel}})+r_{\mathrm{rel}};
\psi(S_{\mathrm{scene}})+r_{\mathrm{scene}}
\right].
\end{equation}
Here, $\psi(\cdot)$ is the cross-modal projector, and $r_{\mathrm{res}}$, $r_{\mathrm{ent}}$, $r_{\mathrm{frame}}$, $r_{\mathrm{rel}}$, and $r_{\mathrm{scene}}$ are lightweight role embeddings. The role embeddings expose the identity of each serialized block to the LLM, while the state semantics are grounded by the evidence sources, induction pathways, and role-preserving objectives introduced in Sec.~\ref{sec:role_preserving_learning}. We provide serialization details in Appendix~\ref{app:state_reader_serialization}.

For unified 3D tasks such as question answering, scene description, and visual grounding, $Z_{\mathrm{state}}$ is used as the shared visual interface for the LLM. Different tasks therefore reason over the same scene-state foundation and share object, spatial, and relational evidence.

For tasks that involve dense mask decoding, we keep the standard $[\mathrm{SEG}]$-based query generation and use the constructed states as a lightweight query condition.
Let $q_{\mathrm{seg}}^{0}$ be the initial segmentation query. The state-conditioned query is computed as
\begin{equation}
q_{\mathrm{seg}}
=
q_{\mathrm{seg}}^{0}
+
\alpha W_{\mathrm{state}}
\left[
\mathrm{Pool}(S_{\mathrm{ent}});
\mathrm{Pool}(S_{\mathrm{frame}});
\mathrm{Pool}(S_{\mathrm{rel}})
\right],
\end{equation}
where $\alpha$ is a learnable scaling factor and $W_{\mathrm{state}}$ maps the pooled states into the query space. This condition allows the mask decoder to access the same object, spatial, and relational summaries used by the LLM, providing additional context for spatially constrained referring expressions.

\subsection{Training Objective}
\label{sec:role_preserving_learning}

The training objective of \method{} combines task supervision with role-preserving supervision:
\begin{equation}
\mathcal{L}
=
\mathcal{L}_{\mathrm{task}}
+
\mathcal{R}_{\mathrm{role}}.
\end{equation}
Here, $\mathcal{L}_{\mathrm{task}}$ includes the standard language modeling loss and the segmentation mask loss. The role-preserving term maintains the structural responsibilities of different states:
\begin{equation}
\mathcal{R}_{\mathrm{role}}
=
\lambda_{\mathrm{ent}}\mathcal{R}_{\mathrm{ent}}
+
\lambda_{\mathrm{frame}}\mathcal{R}_{\mathrm{frame}}
+
\lambda_{\mathrm{rel}}\mathcal{R}_{\mathrm{rel}}.
\end{equation}
We keep the role-preserving objectives compact in the main text and provide the detailed construction of the coverage score $c_g$, the boundary label $y_{\mathrm{bd}}$, and the region label $y_{\mathrm{reg}}$ in Appendix~\ref{app:role_preserving_details}.

\paragraph{Entity semantic preservation.}
Entity states are encouraged to remain consistent with the object-core evidence they read. We construct an entity target by weighted aggregation over entity evidence and align the pooled entity state with this target in the feature space:
\begin{equation}
\mathcal{R}_{\mathrm{ent}}
=
1-
\cos
\left(
g_{\mathrm{ent}}(\mathrm{Pool}(S_{\mathrm{ent}})),
\mathrm{Pool}_{w}(\mathcal{P}_{\mathrm{ent}})
\right).
\end{equation}
Here, $g_{\mathrm{ent}}$ is a lightweight projection layer, and $\mathrm{Pool}_{w}$ denotes score-weighted pooling over entity evidence. This constraint stabilizes the object identity and local semantics preserved by entity states.

\paragraph{Frame coverage preservation.}
Scene-frame states are encouraged to cover diverse boundary and region anchors. Let $A_{\mathrm{frame}}$ be the attention distribution when frame states read frame evidence, and let $\mathcal{G}$ denote the set of boundary and region anchor groups. For each group $g$, we define its coverage as $c_g$. The frame coverage term is
\begin{equation}
\mathcal{R}_{\mathrm{frame}}
=
-
\frac{1}{|\mathcal{G}|}
\sum_{g\in\mathcal{G}}
\log(c_g+\epsilon).
\end{equation}
This term encourages frame states to attend to multiple boundary and region groups, so that the frame representation preserves spatial coverage rather than concentrating on a small subset of anchors.

\paragraph{Relation geometry regularization.}
The relation states model spatial interactions between object evidence and scene-frame references. We apply a coarse geometry-aware regularization that aligns relation states with the boundary and region references derived from entity evidence:
\begin{equation}
\mathcal{R}_{\mathrm{rel}}
=
\mathrm{CE}(\hat{y}_{\mathrm{bd}},y_{\mathrm{bd}})
+
\mathrm{CE}(\hat{y}_{\mathrm{reg}},y_{\mathrm{reg}}).
\end{equation}
Here, $y_{\mathrm{bd}}$ and $y_{\mathrm{reg}}$ are boundary and region labels generated from the weighted center of entity evidence. This term provides lightweight geometric supervision for relation states, encouraging them to carry coarse object-environment references.

\section{Experiments}
\label{sec:experiments}


\subsection{Experimental Setting}
\label{sec:exp_setting}

\paragraph{Datasets and benchmarks.}
We evaluate SceneScaffold on five representative 3D vision-language benchmarks, including ScanRefer~\cite{chen2020scanrefer}, Multi3DRefer~\cite{zhang2023multi3drefer}, ScanQA~\cite{azuma2022scanqa}, SQA3D~\cite{ma2023sqa3d}, and Scan2Cap~\cite{chen2021scan2cap}. ScanRefer evaluates single-target visual grounding, where a model is required to predict a point-level mask according to a natural-language expression. Multi3DRefer extends this setting to zero-target, single-target, and multi-target expressions, and is therefore more suitable for evaluating multi-object disambiguation in complex 3D scenes. ScanQA evaluates question answering over 3D scenes, while SQA3D further introduces situated question answering under embodied scene contexts. Scan2Cap evaluates dense 3D captioning, where the model describes target objects together with their attributes, locations, and surrounding context.
For ScanRefer and Multi3DRefer, we report mean Intersection-over-Union (mIoU). For ScanQA and Scan2Cap, we report CIDEr (C), BLEU-4 (B-4), METEOR (M), and ROUGE-L (R). For SQA3D, we report exact match (EM) and its refined version (EM-R). These benchmarks jointly cover point-level grounding, language-based reasoning, and object description, allowing us to evaluate whether a shared scene-state interface can support heterogeneous 3D scene understanding tasks.

\paragraph{Implementation details.}
For a fair comparison with recent unified 3D LMMs, we follow the unified instruction-tuning protocol of 3D-LLaVA~\cite{deng2025_3dllava}. SceneScaffold uses LLaVA-1.5-7B~\cite{liu2024llava15} as the language-model backbone. The point-cloud encoder adopts a 3D U-Net architecture and is initialized from pretrained weights on large-scale indoor 3D scenes~\cite{deng2025_3dllava}. Given an input point cloud, the point-cloud encoder extracts superpoint-level visual features together with their 3D centroids. SceneScaffold then reorganizes the fixed visual interface into residual detail tokens, entity states, scene-frame states, relation states, and a global scene state.
Unless otherwise specified, the total number of visual tokens inserted into the language model is fixed to $K=100$. The default allocation consists of $75$ residual detail tokens, $8$ entity states, $8$ scene-frame states, $8$ relation states, and $1$ global scene state. During instruction tuning, we freeze the point-cloud encoder and the main body of the LLM, and optimize only the SceneScaffold state-construction modules, cross-modal projection layers, segmentation-query projection layers, and LoRA parameters~\cite{hu2021lora}. We train the model on 8 NVIDIA A800 GPUs using AdamW with a cosine learning-rate schedule. The initial learning rate is set to $2\times10^{-4}$. The batch size, number of epochs, and gradient accumulation steps are set to $2$, $1$, and $8$, respectively. The training objective includes language modeling loss, mask prediction loss, and role-preserving regularization, including entity semantic preservation, frame coverage preservation, and relation geometry regularization. Source code is included in the supplementary material and will be made publicly available upon publication.

\begin{table*}[t]
\caption{\textbf{Overall comparison on five 3D vision-language benchmarks.}
\textit{PC} and \textit{I} denote point cloud and image modality (Mod.), respectively. Specialist models and PC+I are included for reference. 
The best and second-best results among point-cloud-only generalist models are marked in \textbf{bold} and \underline{underline}, respectively. 
}
\label{tab:main_results}
\centering
\setlength{\tabcolsep}{1.5pt}
\renewcommand{\arraystretch}{1.08}
\resizebox{\textwidth}{!}{
\begin{tabular}{l c cc cccc cc cccc}
\toprule
\multirow{2}{*}{Method} & \multirow{2}{*}{Mod.} 
& \multicolumn{1}{c}{ScanRefer} 
& \multicolumn{1}{c}{Multi3D} 
& \multicolumn{4}{c}{ScanQA (val)} 
& \multicolumn{2}{c}{SQA3D (test)} 
& \multicolumn{4}{c}{Scan2Cap (val)} \\
\cmidrule(lr){3-3} 
\cmidrule(lr){4-4} 
\cmidrule(lr){5-8} 
\cmidrule(lr){9-10} 
\cmidrule(lr){11-14}
 & & mIoU$\uparrow$ & mIoU$\uparrow$ 
 & C$\uparrow$ & B-4$\uparrow$ & M$\uparrow$ & R$\uparrow$ 
 & EM$\uparrow$ & EM-R$\uparrow$ 
 & C@.5$\uparrow$ & B-4@.5$\uparrow$ & M@.5$\uparrow$ & R@.5$\uparrow$ \\
\midrule
\rowcolor{groupblue} \multicolumn{14}{l}{\textit{Specialist Models}} \\
ScanQA~\citep{azuma2022scanqa} & PC & -- & -- & 64.9 & 10.1 & 13.1 & 33.3 & 46.6 & -- & -- & -- & -- & -- \\
3D-VLP~\citep{jin2023_3dvlp} & PC & -- & -- & 67.0 & 11.2 & 13.5 & 34.5 & -- & -- & 54.9 & 32.3 & 24.8 & 51.5 \\
3D-VisTA~\citep{zhu2023_3dvista} & PC & -- & -- & 69.6 & 10.4 & 13.9 & 45.7 & 48.5 & -- & 61.6 & 34.1 & 26.8 & 55.0 \\
Scan2Cap~\citep{chen2021scan2cap} & PC & -- & -- & -- & -- & -- & -- & 41.0 & -- & 39.1 & 23.3 & 22.0 & 44.8 \\
MORE~\citep{jiao2022more} & PC & -- & -- & -- & -- & -- & -- & -- & -- & 40.9 & 22.9 & 21.7 & 44.4 \\
SpaCap3D~\citep{wang2022spacap3d} & PC & -- & -- & -- & -- & -- & -- & -- & -- & 44.0 & 25.3 & 22.3 & 45.4 \\
D3Net~\citep{chen2022d3net} & PC & -- & -- & -- & -- & -- & -- & -- & -- & 46.1 & 30.3 & 24.4 & 51.7 \\
UniT3D~\citep{chen2023unit3d} & PC & -- & -- & -- & -- & -- & -- & -- & -- & 46.7 & 27.2 & 21.9 & 46.0 \\
3DJCG~\citep{cai2022_3djcg} & PC & -- & -- & -- & -- & -- & -- & -- & -- & 49.5 & 31.0 & 24.2 & 50.8 \\
Vote2Cap-DETR~\citep{chen2023vote2capdetr} & PC & -- & -- & -- & -- & -- & -- & -- & -- & 61.8 & 34.5 & 26.2 & 54.4 \\
TGNN~\citep{huang2021tgnn} & PC & 27.8 & -- & -- & -- & -- & -- & -- & -- & -- & -- & -- & -- \\
M3DRef-CLIP~\citep{zhang2023multi3drefer} & PC & 35.7 & 32.6 & -- & -- & -- & -- & -- & -- & -- & -- & -- & -- \\
X-RefSeg3D~\citep{qian2024xrefseg3d} & PC & 29.9 & -- & -- & -- & -- & -- & -- & -- & -- & -- & -- & -- \\
3D-STMN~\citep{wu2024_3dstmn} & PC & 39.5 & -- & -- & -- & -- & -- & -- & -- & -- & -- & -- & -- \\
\addlinespace[0.4em]
\rowcolor{groupblue} \multicolumn{14}{l}{\textit{Generalist Models}} \\
3D-LLM~\citep{hong2023_3dllm} & PC+I & -- & -- & 69.4 & 12.0 & 14.5 & 35.7 & -- & -- & -- & -- & -- & -- \\
LEO~\citep{huang2023leo} & PC+I & -- & -- & 101.4 & 13.2 & 20.0 & 49.2 & 50.0 & 52.4 & 72.4 & 38.2 & 27.9 & 58.1 \\
Scene-LLM~\citep{fu2024_scenellm} & PC+I & -- & -- & 80.0 & 11.7 & 15.8 & 35.9 & 53.6 & -- & -- & -- & -- & -- \\
Chat-Scene~\citep{huang2024_chatscene} & PC+I & -- & -- & 87.7 & 14.3 & 18.0 & 41.6 & 54.6 & 57.5 & 77.2 & 36.4 & 28.0 & 58.1 \\
LL3DA~\citep{chen2024_ll3da} & PC & -- & -- & 76.8 & 13.5 & 15.9 & 37.3 & -- & -- & 65.2 & 36.8 & 26.0 & 55.1 \\
Spatial 3D-LLM ~\citep{wang2025spatial} & PC & -- & -- & 82.5 & 13.9 & 16.8 & 39.1 & 46.2 & -- & 72.2 & 34.6 & 23.1 & 54.3 \\
Grounded 3D-LLM~\citep{chen2024_grounded3dllm} & PC & -- & -- & 72.7 & 13.4 & -- & -- & -- & -- & 70.6 & 35.5 & -- & -- \\
LSceneLLM~\citep{zhi2025lscenellm} & PC & -- & -- & 88.2 & -- & 18.0 & 40.8 & -- & -- & -- & -- & -- & -- \\
3D-LLaVA~\citep{deng2025_3dllava} & PC 
& \underline{43.3} & 42.7 
& 92.6 & \textbf{17.1} & 18.4 & 43.1 
& \underline{54.5} & 56.6 
& 78.8 & \underline{36.9} & \underline{27.1} & \underline{57.7} \\
NDTokenizer3D~\citep{tang2025scenes} & PC 
& -- & \underline{46.0} 
& \textbf{98.6} & \underline{17.0} & \textbf{19.4} & \underline{44.9} 
& 54.4 & \underline{57.1} 
& \underline{79.0} & 36.7 & \underline{27.1} & \underline{57.7} \\
\rowcolor{oursblue} \textbf{SceneScaffold (ours)} & PC 
& \textbf{47.0} & \textbf{47.9} 
& \underline{95.8} & 16.8 & \underline{19.0} & \textbf{45.0} 
& \textbf{55.5} & \textbf{58.3} 
& \textbf{79.1} & \textbf{37.1} & \textbf{27.3} & \textbf{57.8} \\
\bottomrule
\end{tabular}
}
\end{table*}

\subsection{Main Results}
\label{sec:main_results}

Table~\ref{tab:main_results} compares SceneScaffold with specialist models and generalist 3D models on five 3D vision-language benchmarks. SceneScaffold uses point clouds as visual input and constructs a role-factored scene-state interface under the same fixed visual-token budget. Specialist models and PC+I methods are included for reference, while the best and second-best results are highlighted among point-cloud-only generalist models.

\paragraph{3D visual grounding.}
On 3D visual grounding, SceneScaffold achieves $47.0$ mIoU on ScanRefer and $47.9$ mIoU on Multi3DRefer, improving over 3D-LLaVA by $3.7$ and $5.2$ mIoU, respectively. It also outperforms NDTokenizer3D on Multi3DRefer by $1.9$ mIoU. These results show that SceneScaffold brings clear gains on localization-oriented tasks, where the model must resolve object identity, spatial references, and multiple similar candidates. This improvement is consistent with the design of SceneScaffold, which organizes the fixed visual interface into entity, frame, and relation states rather than a homogeneous set of object-centric visual tokens.

\paragraph{3D question answering.}
On ScanQA, SceneScaffold obtains $95.8$ CIDEr, $16.8$ BLEU-4, $19.0$ METEOR, and $45.0$ ROUGE-L. Compared with 3D-LLaVA, it improves CIDEr, METEOR, and ROUGE-L by $3.2$, $0.6$, and $1.9$, respectively. On SQA3D, SceneScaffold achieves $55.5$ EM and $58.3$ EM-R, outperforming both 3D-LLaVA and NDTokenizer3D. These results indicate that the constructed scene states also provide useful context for language-based reasoning over 3D scenes.

\paragraph{3D dense captioning.}
On Scan2Cap, SceneScaffold achieves $79.1$ CIDEr@0.5, $37.1$ BLEU-4@0.5, $27.3$ METEOR@0.5, and $57.8$ ROUGE-L@0.5. Compared with 3D-LLaVA, it obtains modest improvements across all four metrics. Overall, SceneScaffold shows the strongest gains on grounding-oriented tasks, while maintaining competitive performance on question answering and dense captioning.

\vspace{-5pt}
\subsection{Diagnostic Evaluation on Relation-Heavy and Ambiguous Scenes} \label{sec:diagnostic}
\vspace{-5pt}

To examine whether SceneScaffold benefits samples that require spatial references and multi-instance disambiguation, we evaluate it on a diagnostic Rel.+Amb. subset. This subset contains samples involving both explicit spatial-reference expressions and multiple same-category candidates; its construction details and statistics are provided in Appendix~\ref{app:diagnostic_subset}.

\begin{table*}[t]
\centering
\caption{
\textbf{Diagnostic evaluation on relation-heavy and spatially ambiguous samples.}
``Overall'' denotes the full evaluation split, while ``Rel.+Amb.'' denotes the subset requiring both spatial-reference understanding and multi-instance disambiguation.
}
\label{tab:diagnostic}
\vspace{0pt}
\setlength{\tabcolsep}{2pt}
\renewcommand{\arraystretch}{1.08}
\begin{tabular}{l cc cc cc cc}
\toprule
\multirow{3}{*}{Method}
& \multicolumn{2}{c}{ScanRefer}
& \multicolumn{2}{c}{Multi3DRefer}
& \multicolumn{2}{c}{ScanQA}
& \multicolumn{2}{c}{Scan2Cap} \\
\cmidrule(lr){2-3}
\cmidrule(lr){4-5}
\cmidrule(lr){6-7}
\cmidrule(lr){8-9}
& Overall 
& Rel.+Amb. 
& Overall 
& Rel.+Amb. 
& Overall 
& Rel.+Amb. 
& Overall 
& Rel.+Amb. \\
& mIoU$\uparrow$
& mIoU$\uparrow$
& mIoU$\uparrow$
& mIoU$\uparrow$
& C$\uparrow$
& C$\uparrow$
& C@0.5$\uparrow$
& C@0.5$\uparrow$ \\
\midrule
3D-LLaVA
& 43.3 & 37.6
& 42.7 & 40.7
& 92.6 & 84.9
& 78.8 & 76.0 \\
\rowcolor{oursblue}
SceneScaffold
& \textbf{47.0} & \textbf{40.6}
& \textbf{47.9} & \textbf{43.5}
& \textbf{95.8} & \textbf{88.6}
& \textbf{79.1} & \textbf{77.9} \\
\bottomrule
\end{tabular}
\end{table*}

As shown in Table~\ref{tab:diagnostic}, the Rel.+Amb. subset is consistently harder than the full split, especially on ScanRefer, ScanQA, and Scan2Cap. SceneScaffold still improves over 3D-LLaVA on all diagnostic metrics, with gains of $3.0$ mIoU on ScanRefer, $2.8$ mIoU on Multi3DRefer, $3.7$ CIDEr on ScanQA, and $1.9$ CIDEr@0.5 on Scan2Cap. These results indicate that the proposed scene-state interface remains effective when spatial references and multiple same-category candidates must be jointly resolved, supporting our motivation of organizing entity, frame, and relation evidence before language reasoning.

\vspace{-5pt}
\subsection{Ablation Study} \label{sec:ablation}
\vspace{-5pt}

We ablate different scene-state components to examine their contributions to unified 3D scene understanding. All variants keep the same final visual-token budget. When one state component is removed, its slots are reallocated to residual detail tokens, so that the comparison focuses on the organization of the visual interface rather than the number of tokens.

\begin{table*}[t]
\centering
\caption{\textbf{Ablation on scene-state components.}
All variants keep the same final visual-token budget. Removed state slots are reallocated to residual detail tokens.}
\label{tab:state_ablation}
\vspace{5pt}
\setlength{\tabcolsep}{1.5pt}
\renewcommand{\arraystretch}{1.08}
\resizebox{\textwidth}{!}{
\begin{tabular}{l cc cccc cc cccc}
\toprule
\multirow{3}{*}{Variant}
& \multicolumn{2}{c}{Grounding}
& \multicolumn{4}{c}{ScanQA}
& \multicolumn{2}{c}{SQA3D}
& \multicolumn{4}{c}{Scan2Cap} \\
\cmidrule(lr){2-3}
\cmidrule(lr){4-7}
\cmidrule(lr){8-9}
\cmidrule(lr){10-13}
& ScanRefer
& Multi3DRefer
& 
& 
& 
& 
& 
& 
& 
& 
& 
&  \\
& mIoU$\uparrow$
& mIoU$\uparrow$
& C$\uparrow$
& B-4$\uparrow$
& M$\uparrow$
& R$\uparrow$
& EM$\uparrow$
& EM-R$\uparrow$
& C@.5$\uparrow$
& B-4@.5$\uparrow$
& M@.5$\uparrow$
& R@.5$\uparrow$ \\
\midrule
w/o $S_{\mathrm{ent}}$
& 45.0 & 46.2
& 91.1 & 14.3 & 18.2 & 43.6
& 52.7 & 56.2
& 77.5 & 36.5 & 27.0 & 57.6 \\
w/o $S_{\mathrm{frame}}$
& 45.9 & 46.2
& 95.3 & 16.7 & 18.9 & 44.8
& 55.3 & 58.3
& 77.9 & 36.9 & 27.1 & 57.8 \\
w/o $S_{\mathrm{rel}}$
& 45.1 & 46.4
& 92.6 & 15.1 & 18.4 & 43.8
& 56.4 & 57.8
& 75.5 & 35.9 & 26.9 & 57.3 \\
w/o $S_{\mathrm{scene}}$
& 45.8 & 46.3
& 93.0 & 15.1 & 18.3 & 44.2
& 54.8 & 57.8
& 76.1 & 35.5 & 27.2 & 57.7 \\
\rowcolor{oursblue}
\textbf{SceneScaffold Full}
& 47.0 & 47.9
& 95.8 & 16.8 & 19.0 & 45.0
& 55.5 & 58.3
& 79.1 & 37.1 & 27.3 & 57.8 \\
\bottomrule
\end{tabular}
}
\end{table*}

Table~\ref{tab:state_ablation} reports the ablation results of different scene-state components under the same final visual-token budget. The full SceneScaffold model achieves the best grounding performance, reaching $47.0$ mIoU on ScanRefer and $47.9$ mIoU on Multi3DRefer. It also obtains the strongest ScanQA results and the most balanced Scan2Cap performance among the ablated variants, showing that the complete state composition provides a more stable visual interface across tasks.
Different components contribute complementary roles. Entity states provide broad object-centric evidence: without $S_{\mathrm{ent}}$, ScanRefer drops from $47.0$ to $45.0$, Multi3DRefer from $47.9$ to $46.2$, and ScanQA CIDEr from $95.8$ to $91.1$. Scene-frame states mainly support spatial anchoring. Removing $S_{\mathrm{frame}}$ keeps several language metrics close to the full model, but consistently weakens grounding performance, which is consistent with its role in preserving boundary and region references. 
Relation states appear to support relational reasoning and spatially grounded description: removing $S_{\mathrm{rel}}$ reduces ScanQA and Scan2Cap performance, with Scan2Cap CIDEr decreasing from $79.1$ to $75.5$.
The global scene state mainly affects language-oriented tasks; without $S_{\mathrm{scene}}$, ScanQA CIDEr decreases to $93.0$ and Scan2Cap CIDEr to $76.1$, reflecting its function as a compact summary of entity, frame, and relation states.

\begin{figure*}[t]
\centering
\includegraphics[width=0.87\textwidth]{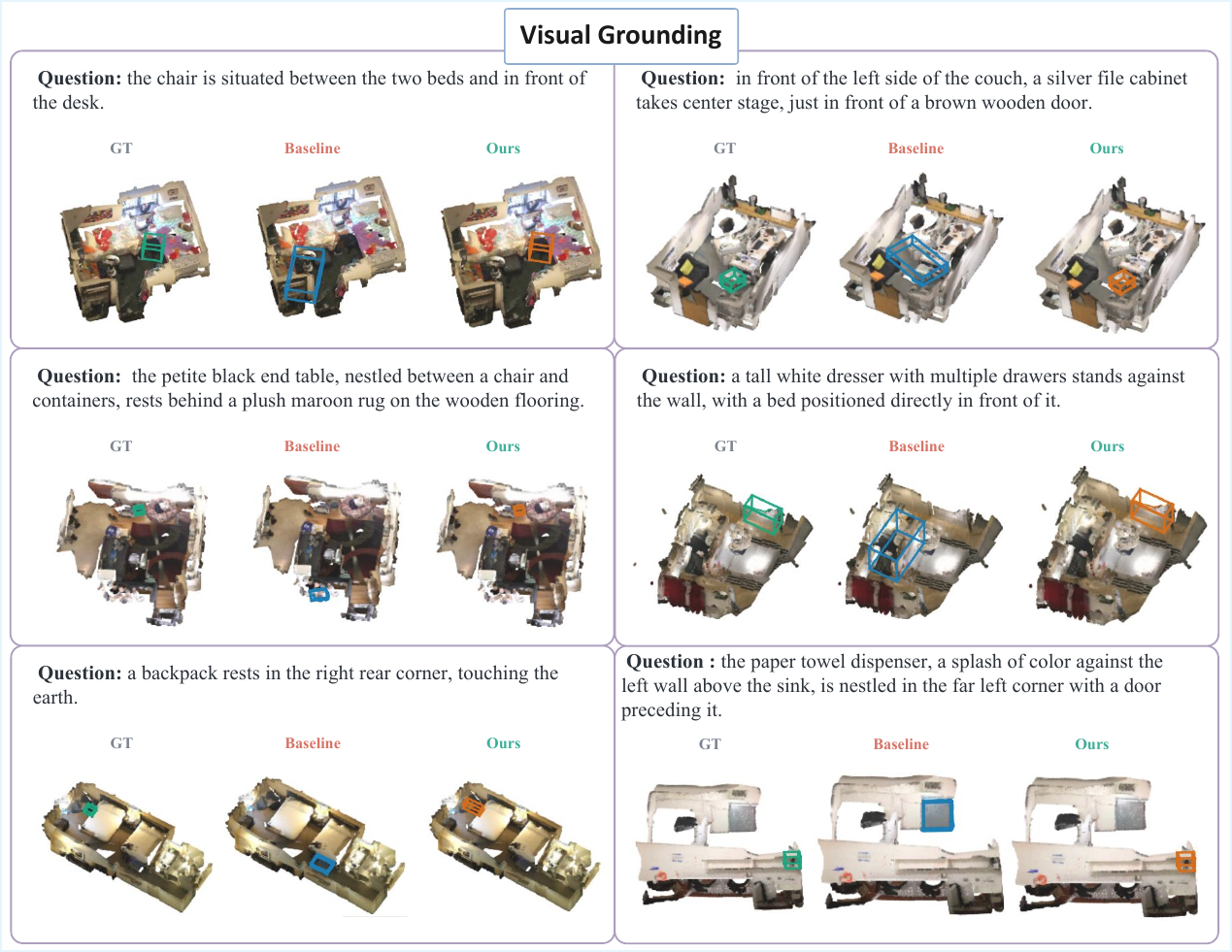}
\caption{
\textbf{Qualitative results on visual grounding.}
SceneScaffold more accurately grounds relation-heavy and spatially ambiguous expressions by organizing entity, frame, and relation states, whereas the baseline tends to select semantically similar but spatially incorrect instances.
}
\label{fig:grounding_qualitative}
\vspace{-10pt}
\end{figure*}

Overall, replacing any state component with additional residual detail tokens cannot recover the full model, indicating that SceneScaffold benefits from role-factored scene-state organization rather than merely from local detail preservation. Figure~\ref{fig:grounding_qualitative} further illustrates this behavior on relation-heavy and spatially ambiguous samples, complementing the diagnostic results in Table~\ref{tab:diagnostic}: the selected cases contain multiple same-category candidates and explicit spatial references, such as objects located between beds, near a wall, in front of another object, or in a room corner. The baseline often identifies the correct object category but drifts to a spatially inconsistent instance, whereas SceneScaffold more reliably grounds the target satisfying the relation constraint. Additional qualitative results on question answering and dense captioning are provided in Appendix~\ref{app:additional_qualitative}. Appendix~\ref{app:loss_ablation} additionally shows that removing role-preserving objectives weakens performance while keeping the architecture and token budget unchanged, supporting their role in stabilizing the intended state functions.

\vspace{-5pt}
\section{Conclusion} \label{sec:conclusion}
\vspace{-5pt}

In this paper, we revisit the visual bottleneck in unified 3D large multimodal models from the perspective of modal role allocation. Instead of treating the bottleneck as a passive feature compressor that selects a homogeneous set of visual fragments, we argue that the visual side should organize a compact 3D scene state before language reasoning. To this end, we propose SceneScaffold, an active scene-state construction framework that reallocates a fixed visual-token budget into residual detail tokens, entity states, scene-frame states, relation states, and a global scene summary. These states are grounded in distinct evidence sources and induction pathways, allowing object semantics, spatial references, and interaction cues to be exposed to the language model in a structured form.
Experiments across grounding, question answering, and dense captioning show that SceneScaffold improves unified 3D scene understanding, with larger gains on relation-heavy and spatially ambiguous samples. These results suggest that active scene-state construction is an effective direction for building more reliable 3D-LMM interfaces under limited visual budgets.

\begin{ack}
This work was supported by Beijing Natural Science Foundation (No.~QG26011), National Natural Science Foundation of China (No.~62606510, No.~62476264, No.~62406312, and No.~62676389), and the Youth Key Project of the Chinese Academy of Sciences (GFQN-2026-34).
\end{ack}

{
\small
\bibliography{reference}
}
\newpage

\appendix

\makeatletter
\@addtoreset{figure}{section}
\@addtoreset{table}{section}
\makeatother

\renewcommand{\thefigure}{\thesection.\arabic{figure}}
\renewcommand{\thetable}{\thesection.\arabic{table}}

{\centering\Large\bfseries
Appendix for \textit{
SceneScaffold: Active Scene-State Construction for Unified 3D Scene Understanding
}\par}
\vspace{1em}

\section{Implementation Details of Scene-State Construction and Learning}
\label{app:role_preserving_details}

This section provides implementation details for state reader construction, serialization, role-grounded evidence construction, and role-preserving learning in \method{}. In the main paper, we present compact forms of entity semantic preservation, frame coverage preservation, and relation geometry regularization. We also summarize the default hyperparameter settings used in all main experiments. Here, we further specify the construction of the weighted entity target, frame group coverage $c_g$, boundary label $y_{\mathrm{bd}}$, and region label $y_{\mathrm{reg}}$.

\subsection{State Reader, Serialization, and Hyperparameter Settings}
\label{app:state_reader_serialization}

We instantiate each state induction operator $\Phi$ as a one-layer cross-attention reader with residual connection and layer normalization. Given learnable role queries $Q$ and evidence tokens $P$, the reader is implemented as
\begin{equation}
\Phi(Q,P)
=
\mathrm{LN}
\left(
Q+\mathrm{MHA}(Q,P,P)
\right),
\end{equation}
where $\mathrm{MHA}(\cdot)$ denotes multi-head cross-attention. Entity, frame, and relation states use independent query parameters and independent reader blocks. The reader uses four attention heads, attention dropout $0.0$, and temperature $1.0$.

All serialized visual tokens are mapped by the same cross-modal projector $\psi(\cdot)$. The serialized order is fixed for all scenes:
\begin{equation}
[
Z_{\mathrm{res}},
S_{\mathrm{ent}},
S_{\mathrm{frame}},
S_{\mathrm{rel}},
S_{\mathrm{scene}}
].
\end{equation}
After projection, we add one learnable role embedding to each block type, corresponding to residual, entity, frame, relation, and scene-summary tokens. These role embeddings are initialized to zero and learned during training. We do not introduce an additional role-specific projector; all state blocks share the same visual-to-language projector before role embeddings are added.

In our implementation, the total visual budget is $K=100$. We use $8$ entity state slots, $8$ scene-frame state slots, $8$ relation state slots, and one global scene summary token. Unless otherwise specified, the residual object buffer uses the remaining budget, resulting in $75$ residual object tokens. For role-grounded evidence construction, the entity primitive budget is $128$. The scene-frame evidence uses four boundary anchors and a $3\times3$ region grid, with a boundary-band ratio of $0.15$ and a minimum of four superpoints required for a valid anchor. The frame evidence budget is therefore determined by the valid boundary and occupied region anchors, up to the configured frame-anchor budget. The state-conditioned segmentation query uses a learnable scale initialized to $0.0$. The role-preserving loss weights are set to $0.05$ for entity semantic preservation, frame coverage preservation, relation boundary regularization, and relation region regularization. The residual object dropout rate is set to $0.2$ during training and disabled during inference.

\subsection{Entity Target for Semantic Preservation}
\label{app:entity_target}

Given superpoint features $H=\{h_j\}_{j=1}^{M}$, the entity scoring head predicts an entity score for each superpoint:
\begin{equation}
a_j=f_{\mathrm{ent}}(h_j).
\end{equation}
The scoring head $f_{\mathrm{ent}}$ is implemented as a lightweight MLP:
\begin{equation}
f_{\mathrm{ent}}(h_j)
=
W_2
\sigma
\left(
W_1\mathrm{LN}(h_j)
\right),
\end{equation}
where $\mathrm{LN}(\cdot)$ is layer normalization and $\sigma(\cdot)$ is GELU. The output is a scalar entity score for each superpoint.
We select the top-scored superpoints as entity evidence:
\begin{equation}
\mathcal{I}_{\mathrm{ent}}
=
\mathrm{TopK}(a,K_{\mathrm{ent}}),
\quad
\mathcal{P}_{\mathrm{ent}}
=
\{h_j\mid j\in\mathcal{I}_{\mathrm{ent}}\}.
\end{equation}
We use hard TopK selection in this step. Gradients are propagated through the selected feature paths, while the discrete index selection itself is not relaxed.
The selected entity scores are normalized as
\begin{equation}
w_j
=
\frac{\exp(a_j)}
{\sum_{k\in\mathcal{I}_{\mathrm{ent}}}\exp(a_k)},
\quad
j\in\mathcal{I}_{\mathrm{ent}}.
\end{equation}
The score-weighted entity target is
\begin{equation}
t_{\mathrm{ent}}
=
\mathrm{Pool}_{w}(\mathcal{P}_{\mathrm{ent}})
=
\sum_{j\in\mathcal{I}_{\mathrm{ent}}}w_jh_j.
\end{equation}
The pooled entity state is
\begin{equation}
\bar{S}_{\mathrm{ent}}
=
\mathrm{Pool}(S_{\mathrm{ent}}).
\end{equation}
The entity semantic preservation term used in the main paper is implemented as
\begin{equation}
\mathcal{R}_{\mathrm{ent}}
=
1-
\cos
\left(
g_{\mathrm{ent}}(\bar{S}_{\mathrm{ent}}),
t_{\mathrm{ent}}
\right),
\end{equation}
where $g_{\mathrm{ent}}$ is a lightweight projection layer. This term aligns entity states with the object-core target aggregated from entity evidence.

\subsection{Construction of Frame Anchors}
\label{app:frame_anchor_construction}

Scene-frame evidence consists of boundary anchors and region anchors. For each superpoint centroid
\begin{equation}
u_j=(x_j,y_j,z_j),
\end{equation}
we compute the horizontal scene extent:
\begin{equation}
x_{\min},x_{\max},y_{\min},y_{\max}.
\end{equation}

\paragraph{Boundary anchors.}
Given a boundary-band ratio $\rho$, we define four horizontal boundary bands:
\begin{equation}
\mathcal{B}_{\mathrm{left}}
=
\{j\mid x_j\le x_{\min}+\rho(x_{\max}-x_{\min})\},
\end{equation}
\begin{equation}
\mathcal{B}_{\mathrm{right}}
=
\{j\mid x_j\ge x_{\max}-\rho(x_{\max}-x_{\min})\},
\end{equation}
\begin{equation}
\mathcal{B}_{\mathrm{front}}
=
\{j\mid y_j\le y_{\min}+\rho(y_{\max}-y_{\min})\},
\end{equation}
\begin{equation}
\mathcal{B}_{\mathrm{back}}
=
\{j\mid y_j\ge y_{\max}-\rho(y_{\max}-y_{\min})\}.
\end{equation}
For each non-empty boundary band $q\in\{\mathrm{left},\mathrm{right},\mathrm{front},\mathrm{back}\}$, its boundary anchor is obtained by average pooling:
\begin{equation}
a_q^{\mathrm{bd}}
=
\frac{1}{|\mathcal{B}_q|}
\sum_{j\in\mathcal{B}_q}h_j.
\end{equation}
All boundary anchors form
\begin{equation}
\mathcal{A}_{\mathrm{bd}}
=
\{a_q^{\mathrm{bd}}\}.
\end{equation}

\paragraph{Region anchors.}
For region anchors, we divide the horizontal plane into a $G_x\times G_y$ grid. We first normalize the horizontal coordinates:
\begin{equation}
\tilde{x}_j=
\frac{x_j-x_{\min}}{x_{\max}-x_{\min}+\epsilon},
\quad
\tilde{y}_j=
\frac{y_j-y_{\min}}{y_{\max}-y_{\min}+\epsilon}.
\end{equation}
The region index of each superpoint is
\begin{equation}
r_j
=
\left\lfloor G_x\tilde{x}_j \right\rfloor
+
G_x
\left\lfloor G_y\tilde{y}_j \right\rfloor.
\end{equation}
For each occupied region cell $r$, we define
\begin{equation}
\mathcal{B}_{r}^{\mathrm{reg}}
=
\{j\mid r_j=r\}.
\end{equation}
The corresponding region anchor is
\begin{equation}
a_r^{\mathrm{reg}}
=
\frac{1}{|\mathcal{B}_{r}^{\mathrm{reg}}|}
\sum_{j\in\mathcal{B}_{r}^{\mathrm{reg}}}h_j.
\end{equation}
All region anchors form
\begin{equation}
\mathcal{A}_{\mathrm{reg}}
=
\{a_r^{\mathrm{reg}}\}.
\end{equation}
The final scene-frame evidence is
\begin{equation}
\mathcal{P}_{\mathrm{frame}}
=
\mathcal{A}_{\mathrm{bd}}
\cup
\mathcal{A}_{\mathrm{reg}}.
\end{equation}
In implementation, boundary anchors are constructed first, and the remaining frame-evidence budget is assigned to occupied region anchors. When the number of occupied region cells exceeds the remaining budget, we keep the cells with larger superpoint counts. With the default setting, we use four boundary anchors and a $3\times3$ region grid; only anchors satisfying the minimum-point requirement are used for state induction.

\subsection{Frame Coverage Term}
\label{app:frame_coverage}

The frame coverage term encourages $S_{\mathrm{frame}}$ to cover diverse boundary and region anchors. Let
\begin{equation}
A_{\mathrm{frame}}
\in
\mathbb{R}^{M_{\mathrm{frame}}\times |\mathcal{P}_{\mathrm{frame}}|}
\end{equation}
be the cross-attention map produced when frame states read frame evidence. $A_{\mathrm{frame}}^{m,j}$ denotes the attention weight from the $m$-th frame state slot to the $j$-th frame primitive.

Each frame primitive is assigned a group id. Boundary anchors are grouped by their directions, i.e., left, right, front, and back. Region anchors are grouped by their spatial grid indices. We denote the set of all valid groups as
\begin{equation}
\mathcal{G}
=
\mathcal{G}_{\mathrm{bd}}
\cup
\mathcal{G}_{\mathrm{reg}}.
\end{equation}
For a group $g\in\mathcal{G}$, its coverage is defined as
\begin{equation}
c_g
=
\max_{m}
\sum_{j:g_j=g}
A_{\mathrm{frame}}^{m,j},
\end{equation}
where $g_j$ is the group id of the $j$-th frame primitive. This definition considers a group as covered if at least one frame state slot assigns sufficient attention to primitives in that group.

The frame coverage regularizer is
\begin{equation}
\mathcal{R}_{\mathrm{frame}}
=
-
\frac{1}{|\mathcal{G}|}
\sum_{g\in\mathcal{G}}
\log(c_g+\epsilon).
\end{equation}
This term increases the probability that different spatial anchors are read by frame states, allowing $S_{\mathrm{frame}}$ to maintain boundary and region coverage.

\subsection{Geometry Labels for Relation Regularization}
\label{app:relation_geometry_labels}

Relation geometry regularization uses the center of entity evidence to generate coarse boundary and region labels. We first compute the weighted entity center:
\begin{equation}
\mu_{\mathrm{ent}}
=
\sum_{j\in\mathcal{I}_{\mathrm{ent}}}w_ju_j,
\end{equation}
where
\begin{equation}
\mu_{\mathrm{ent}}=(\mu_x,\mu_y,\mu_z).
\end{equation}

\paragraph{Boundary label.}
We compute the distances from the entity center to four horizontal boundaries:
\begin{equation}
d_{\mathrm{left}}=\mu_x-x_{\min},
\quad
d_{\mathrm{right}}=x_{\max}-\mu_x,
\end{equation}
\begin{equation}
d_{\mathrm{front}}=\mu_y-y_{\min},
\quad
d_{\mathrm{back}}=y_{\max}-\mu_y.
\end{equation}
The boundary label is defined as the nearest boundary direction:
\begin{equation}
y_{\mathrm{bd}}
=
\arg\min
\{
d_{\mathrm{left}},
d_{\mathrm{right}},
d_{\mathrm{front}},
d_{\mathrm{back}}
\}.
\end{equation}
This label provides a coarse directional reference from the entity evidence to the scene boundary.

\paragraph{Region label.}
The region label is generated according to the grid cell containing the entity center. We normalize the entity center coordinates as
\begin{equation}
\tilde{\mu}_x=
\frac{\mu_x-x_{\min}}{x_{\max}-x_{\min}+\epsilon},
\quad
\tilde{\mu}_y=
\frac{\mu_y-y_{\min}}{y_{\max}-y_{\min}+\epsilon}.
\end{equation}
The region label is
\begin{equation}
y_{\mathrm{reg}}
=
\left\lfloor G_x\tilde{\mu}_x \right\rfloor
+
G_x
\left\lfloor G_y\tilde{\mu}_y \right\rfloor.
\end{equation}
In implementation, $\tilde{\mu}_x$ and $\tilde{\mu}_y$ are clamped to valid ranges to avoid numerical boundary issues.

\subsection{Relation Geometry Regularization}
\label{app:relation_geo_regularization}

Relation states are induced from the interaction between entity states and scene-frame states. To encourage them to carry coarse spatial references between object evidence and spatial anchors, we apply boundary and region classification losses to the pooled relation state:
\begin{equation}
\bar{S}_{\mathrm{rel}}
=
\mathrm{Pool}(S_{\mathrm{rel}}).
\end{equation}
The boundary and region predictions are
\begin{equation}
\hat{y}_{\mathrm{bd}}
=
g_{\mathrm{bd}}(\bar{S}_{\mathrm{rel}}),
\quad
\hat{y}_{\mathrm{reg}}
=
g_{\mathrm{reg}}(\bar{S}_{\mathrm{rel}}),
\end{equation}
where $g_{\mathrm{bd}}$ and $g_{\mathrm{reg}}$ are lightweight classification heads. The relation geometry regularizer is
\begin{equation}
\mathcal{R}_{\mathrm{rel}}
=
\mathrm{CE}(\hat{y}_{\mathrm{bd}},y_{\mathrm{bd}})
+
\mathrm{CE}(\hat{y}_{\mathrm{reg}},y_{\mathrm{reg}}).
\end{equation}
This is a coarse geometry-aware regularization. It encourages $S_{\mathrm{rel}}$ to encode relative references between object evidence, scene boundaries, and occupied regions.

\subsection{Residual Object Dropout}
\label{app:residual_dropout}

\method{} keeps $Z_{\mathrm{res}}$ to preserve the local detail capability of the original visual connector. During training, we apply lightweight dropout to the residual object buffer:
\begin{equation}
\tilde{Z}_{\mathrm{res}}
=
m\odot Z_{\mathrm{res}},
\quad
m_i\sim \mathrm{Bernoulli}(1-p).
\end{equation}
$\tilde{Z}_{\mathrm{res}}$ replaces $Z_{\mathrm{res}}$ during serialization in training. This serves as a state-usage regularizer, encouraging the model to use entity, frame, relation, and global states while preserving local object details. Dropout is disabled during inference.

\subsection{Overall Training Objective}
\label{app:overall_training}

The main paper writes the training objective as
\begin{equation}
\mathcal{L}
=
\mathcal{L}_{\mathrm{task}}
+
\mathcal{R}_{\mathrm{role}},
\end{equation}
where
\begin{equation}
\mathcal{R}_{\mathrm{role}}
=
\lambda_{\mathrm{ent}}\mathcal{R}_{\mathrm{ent}}
+
\lambda_{\mathrm{frame}}\mathcal{R}_{\mathrm{frame}}
+
\lambda_{\mathrm{rel}}\mathcal{R}_{\mathrm{rel}}.
\end{equation}
In implementation, the task loss contains a language modeling loss and a segmentation loss. The complete objective can be written as
\begin{equation}
\mathcal{L}
=
\mathcal{L}_{\mathrm{lm}}
+
\lambda_{\mathrm{seg}}\mathcal{L}_{\mathrm{seg}}
+
\lambda_{\mathrm{ent}}\mathcal{R}_{\mathrm{ent}}
+
\lambda_{\mathrm{frame}}\mathcal{R}_{\mathrm{frame}}
+
\lambda_{\mathrm{rel}}\mathcal{R}_{\mathrm{rel}}.
\end{equation}
For visual grounding, the segmentation loss includes BCE, Dice, and decoder auxiliary losses:
\begin{equation}
\mathcal{L}_{\mathrm{seg}}
=
\lambda_{\mathrm{bce}}\mathcal{L}_{\mathrm{BCE}}
+
\lambda_{\mathrm{dice}}\mathcal{L}_{\mathrm{Dice}}
+
\mathcal{L}_{\mathrm{aux}}^{\mathrm{dec}}.
\end{equation}
The overall objective combines task supervision, role-preserving supervision, and residual-object dropout as a lightweight state-usage regularizer.

\section{Additional Experimental Details}
\label{app:exp_details}

\subsection{Diagnostic Subset Construction}
\label{app:diagnostic_subset}

We construct a relation-heavy and spatially ambiguous diagnostic subset to analyze whether SceneScaffold improves samples that require both spatial-reference understanding and multi-instance disambiguation. We first select samples whose expressions or questions contain explicit spatial or scene-reference words, including \emph{left}, \emph{right}, \emph{front}, \emph{behind}, \emph{near}, \emph{next to}, \emph{beside}, \emph{between}, \emph{under}, \emph{below}, \emph{above}, \emph{on top of}, \emph{against}, \emph{wall}, \emph{window}, \emph{door}, \emph{floor}, \emph{corner}, \emph{closest}, \emph{second}, and \emph{third}. We then keep samples where multiple same-category candidates appear in the same scene. Samples satisfying both criteria form the Rel.+Amb. subset. The resulting subset statistics are summarized in Table~\ref{tab:appendix_stats_budget} (left).

\subsection{State Ablation Budget Allocation}
\label{app:ablation_budget}

All state-component ablations in Table~\ref{tab:state_ablation} keep the total visual-token budget fixed at $K=100$. The full SceneScaffold interface uses
\[
75D+8E+8F+8R+1G=100,
\]
where $D$ denotes residual detail tokens, and $E$, $F$, $R$, and $G$ denote entity states, frame states, relation states, and the global scene state, respectively. When a state component is removed, its budget is reallocated to residual detail tokens. The detailed allocation is reported in Table~\ref{tab:appendix_stats_budget} (right).

\begin{table*}[t]
\centering
\caption{
\textbf{Additional experimental statistics.}
Left: statistics of the diagnostic Rel.+Amb. subsets. Right: token-budget allocation for state-component ablations.
}
\label{tab:appendix_stats_budget}
\small
\setlength{\tabcolsep}{4.0pt}
\renewcommand{\arraystretch}{1.08}
\begin{minipage}[t]{0.36\textwidth}
\centering
\textbf{Diagnostic subset statistics}\\[0.4em]
\begin{tabular}{lcc}
\toprule
Dataset & Total & Rel.+Amb. \\
\midrule
ScanRefer & 9508 & 6329 \\
Multi3DRefer & 11120 & 6235 \\
ScanQA & 4675 & 2272 \\
Scan2Cap & 2007 & 1631 \\
\bottomrule
\end{tabular}
\end{minipage}
\hfill
\begin{minipage}[t]{0.60\textwidth}
\centering
\textbf{Token-budget allocation}\\[0.4em]
\begin{tabular}{lccccc}
\toprule
Variant & Detail & Entity & Frame & Relation & Scene \\
\midrule
Full & 75 & 8 & 8 & 8 & 1 \\
w/o $S_{\mathrm{ent}}$ & 83 & 0 & 8 & 8 & 1 \\
w/o $S_{\mathrm{frame}}$ & 83 & 8 & 0 & 8 & 1 \\
w/o $S_{\mathrm{rel}}$ & 83 & 8 & 8 & 0 & 1 \\
w/o $S_{\mathrm{scene}}$ & 76 & 8 & 8 & 8 & 0 \\
\bottomrule
\end{tabular}
\end{minipage}
\end{table*}

\subsection{Control Baselines for State Construction}
\label{app:state_construction_controls}

We further evaluate several same-budget control baselines to analyze which factors are responsible for the role-factored scene-state interface. Different from the state-component ablation in Table~\ref{tab:state_ablation}, these variants keep the state-interface scale unchanged and modify how the states are constructed.

\textbf{Generic query states} replace the role-factored entity, frame, and relation states with homogeneous learned query states under the same state-slot budget. This baseline tests whether a generic learned token set is sufficient. \textbf{Shared-evidence role states} keep separate state readers and serialized role blocks, but use the same evidence source for different states, testing whether role-specific evidence grounding is necessary. \textbf{Object-centric frame evidence} replaces geometry-derived frame evidence with object-centric evidence, examining whether scene-frame states require boundary and region anchors. \textbf{Relation from primitive evidence} induces relation states directly from primitive evidence instead of the interaction between entity and frame states, testing the role of the entity-frame induction pathway.

\begin{table*}[t]
\centering
\caption{\textbf{Control baselines for role-factored state construction.}
All variants keep the same final visual-token budget. Each baseline modifies one construction factor while preserving the overall state-interface scale.}
\label{tab:state_construction_controls}
\setlength{\tabcolsep}{1.5pt}
\renewcommand{\arraystretch}{1.08}
\resizebox{\textwidth}{!}{
\begin{tabular}{l cc cccc cc cccc}
\toprule
\multirow{2}{*}{Variant}
& \multicolumn{2}{c}{Grounding}
& \multicolumn{4}{c}{ScanQA}
& \multicolumn{2}{c}{SQA3D}
& \multicolumn{4}{c}{Scan2Cap} \\
\cmidrule(lr){2-3}
\cmidrule(lr){4-7}
\cmidrule(lr){8-9}
\cmidrule(lr){10-13}
& ScanRefer mIoU$\uparrow$
& Multi3DRefer mIoU$\uparrow$
& C$\uparrow$
& B-4$\uparrow$
& M$\uparrow$
& R$\uparrow$
& EM$\uparrow$
& EM-R$\uparrow$
& C@.5$\uparrow$
& B-4@.5$\uparrow$
& M@.5$\uparrow$
& R@.5$\uparrow$ \\
\midrule
Generic query states
& 45.7 & 46.5
& 93.6 & 14.9 & 18.5 & 44.4
& 53.4 & 56.1
& 77.7 & 36.7 & 27.0 & 57.5 \\
Shared-evidence role states
& 45.9 & 46.4
& 92.4 & 14.3 & 18.4 & 44.1
& 53.7 & 56.3
& 77.6 & 36.2 & 27.1 & 57.5 \\
Object-centric frame evidence
& 45.8 & 46.7
& 94.2 & 15.2 & 18.7 & 44.7
& 54.1 & 56.7
& 77.5 & 36.1 & 27.0 & 57.3 \\
Relation from primitive evidence
& 45.7 & 46.6
& 92.2 & 14.6 & 18.2 & 44.1
& 54.0 & 56.6
& 77.9 & 36.3 & 27.1 & 57.6 \\
\rowcolor{oursblue}
\textbf{SceneScaffold Full}
& 47.0 & 47.9
& 95.8 & 16.8 & 19.0 & 45.0
& 55.5 & 58.3
& 79.1 & 37.1 & 27.3 & 57.8 \\
\bottomrule
\end{tabular}
}
\end{table*}

Table~\ref{tab:state_construction_controls} shows that all construction-control variants underperform the full SceneScaffold model. Generic query states use the same state-slot budget but obtain lower grounding and QA results, indicating that a homogeneous learned state set provides weaker scene evidence than the role-factored interface. Shared-evidence role states also lag behind the full model, showing that separate role blocks benefit from evidence sources matched to their intended functions. Replacing geometry-derived frame evidence with object-centric evidence reduces grounding performance, which supports the use of boundary and region anchors for spatial reference preservation. Inducing relation states directly from primitive evidence weakens both grounding and language-oriented metrics, suggesting that relation states are more effective when formed through the interaction between entity and frame states. These results complement the state-component ablation in Table~\ref{tab:state_ablation} by showing that SceneScaffold benefits from the coupled design of role-specific evidence, state-induction pathways, and fixed-budget serialization.

\subsection{Functional State Intervention Analysis}
\label{app:state_intervention}

To complement the retrained state-component ablation, we conduct a functional state intervention analysis on a trained full SceneScaffold model. During inference, we keep the residual detail tokens, language input, model parameters, and all other states unchanged, and replace one type of state with the corresponding state sampled from another scene:
\[
S_{\tau}\leftarrow S_{\tau}^{\mathrm{other}},
\quad
\tau\in\{\mathrm{ent},\mathrm{frame},\mathrm{rel},\mathrm{scene}\}.
\]
Compared with zero masking, cross-scene replacement uses valid states produced by the same model, and therefore reduces the influence of obvious out-of-distribution perturbations.

We evaluate this intervention on the Rel.+Amb. diagnostic subsets defined in Appendix~\ref{app:diagnostic_subset}. These subsets contain samples that require both explicit spatial-reference understanding and multi-instance disambiguation. Therefore, they provide a more focused stress test for examining whether the constructed states are functionally used in difficult 3D grounding and reasoning cases. We do not introduce additional entity-centric, frame-reference, or object-relation subsets; all results in this analysis are evaluated on the same Rel.+Amb. subsets.

\begin{table*}[t]
\centering
\caption{\textbf{Functional state intervention analysis on Rel.+Amb. subsets.}
We replace one type of state with the corresponding state from another scene during inference. All results are evaluated on the diagnostic Rel.+Amb. subsets defined in Appendix~\ref{app:diagnostic_subset}.}
\label{tab:state_intervention}
\setlength{\tabcolsep}{4.5pt}
\renewcommand{\arraystretch}{1.08}
\begin{tabular}{lcccc}
\toprule
Intervention
& ScanRefer mIoU$\uparrow$
& Multi3DRefer mIoU$\uparrow$
& ScanQA C$\uparrow$
& Scan2Cap C@.5$\uparrow$ \\
\midrule
Full
& 40.6 & 43.5 & 88.6 & 77.9 \\
Shuffle $S_{\mathrm{ent}}$
& 39.1 & 42.0 & 87.4 & 76.7 \\
Shuffle $S_{\mathrm{frame}}$
& 38.4 & 41.3 & 87.2 & 76.5 \\
Shuffle $S_{\mathrm{rel}}$
& 38.6 & 41.6 & 86.9 & 76.2 \\
Shuffle $S_{\mathrm{scene}}$
& 39.8 & 42.8 & 86.6 & 76.4 \\
Shuffle all states
& 37.8 & 40.8 & 85.0 & 76.0 \\
\bottomrule
\end{tabular}
\end{table*}

Table~\ref{tab:state_intervention} reports functional state intervention results on the Rel.+Amb. subsets. The Full row is identical to the Rel.+Amb. performance of SceneScaffold in Table~\ref{tab:diagnostic}. Replacing individual state types with states from another scene consistently degrades performance, indicating that the constructed states are functionally used by the trained model.
Shuffling $S_{\mathrm{frame}}$ causes clear drops on ScanRefer and Multi3DRefer, suggesting that frame states provide spatial-reference cues for relation-heavy and multi-instance disambiguation. Shuffling $S_{\mathrm{rel}}$ also weakens grounding and captioning performance, indicating that relation states contribute to object--object and object--environment relation modeling. Shuffling $S_{\mathrm{scene}}$ has a relatively smaller effect on grounding but a larger effect on ScanQA and Scan2Cap, which is consistent with its role as a compact global scene summary for language reasoning. Shuffling all states leads to the largest degradation and approaches the baseline performance on the Rel.+Amb. subsets, while still retaining residual detail tokens.

This analysis is not intended as a strict causal explanation of the LLM's internal reasoning process. Instead, it provides functional evidence that the constructed states are not interchangeable latent tokens: replacing different state types affects the difficult Rel.+Amb. subset in different ways.

\subsection{Qualitative Results}
\label{app:additional_qualitative}

\begin{figure*}[t]
\centering
\includegraphics[width=\textwidth]{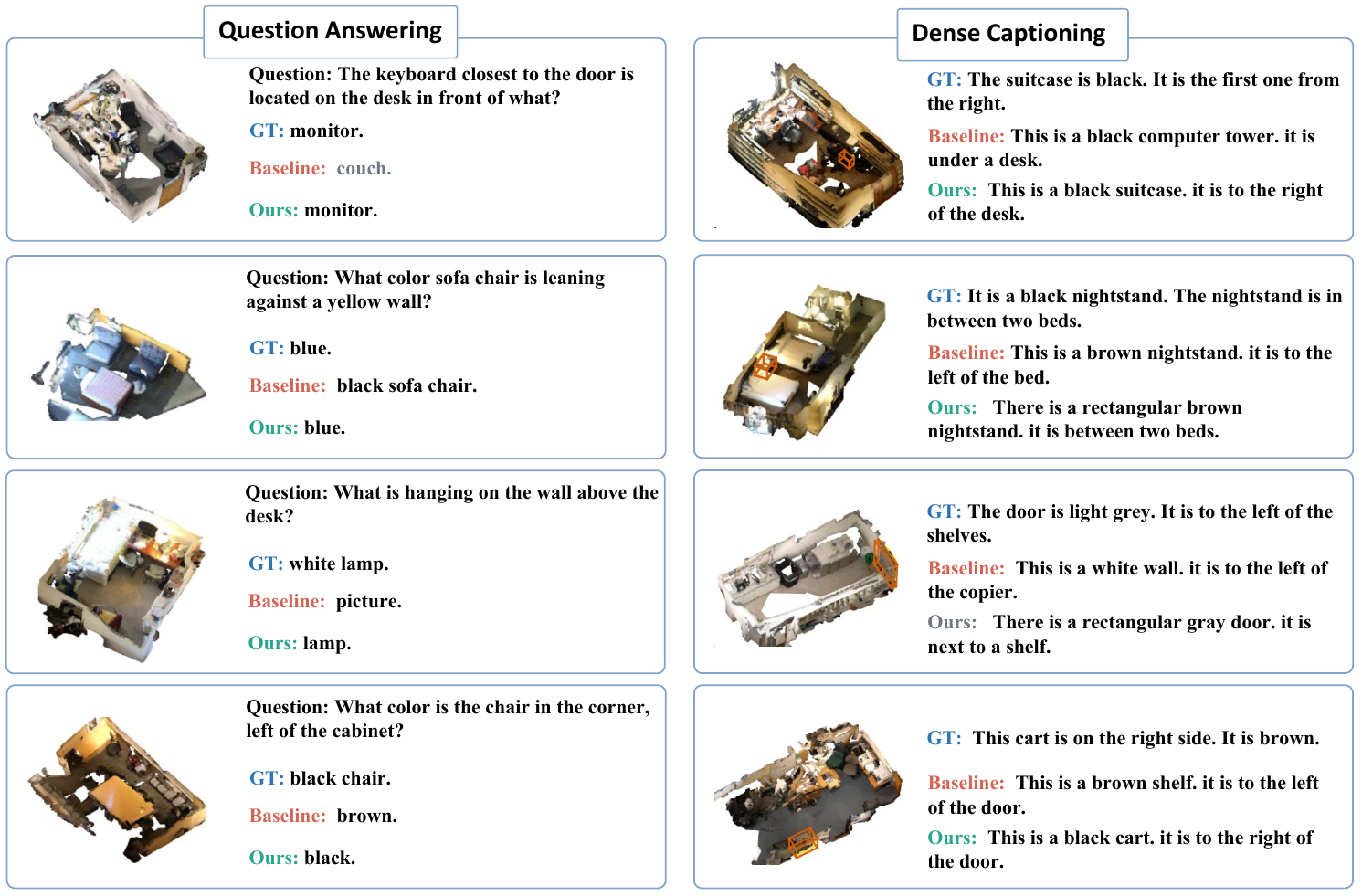}
\caption{
\textbf{Qualitative results on question answering and dense captioning.}
SceneScaffold produces responses that better match the target object identity and spatial context, while the baseline often confuses nearby objects or misses relation cues.
}
\label{fig:qa_caption_qualitative}
\end{figure*}

We provide additional qualitative comparisons in Figure~\ref{fig:qa_caption_qualitative}, complementing the visual grounding cases in Figure~\ref{fig:grounding_qualitative} of the main text. 
Figure~\ref{fig:qa_caption_qualitative} shows examples on question answering and dense captioning. 
Compared with the baseline, SceneScaffold produces answers and descriptions that better preserve spatial references, object identities, and scene context. 
These examples suggest that the constructed scene states are useful not only for localization-heavy tasks, but also for language generation tasks requiring spatial reasoning.

\subsection{Additional Loss Ablation}
\label{app:loss_ablation}

We additionally ablate the role-preserving objectives used to stabilize different scene states. 
Different from the scene-state component ablation in the main text, this analysis keeps the model architecture and the visual-token budget unchanged, and only removes the corresponding training objective. 
Therefore, this experiment does not test whether a state component is necessary; instead, it examines whether the intended role of each existing state can be better maintained by its role-preserving objective.

Specifically, we consider entity semantic preservation $\mathcal{R}_{\mathrm{ent}}$, frame coverage preservation $\mathcal{R}_{\mathrm{frame}}$, and relation geometry regularization $\mathcal{R}_{\mathrm{rel}}$. 
The entity objective encourages entity states to remain aligned with object-core evidence. 
The frame objective encourages frame states to cover diverse boundary and region anchors. 
The relation objective provides coarse geometry-aware supervision for relation states.

\begin{table*}[t]
\centering
\caption{\textbf{Ablation on role-preserving objectives.}
All variants keep the same model architecture and visual-token budget. Only the corresponding role-preserving objective is removed.}
\label{tab:loss_ablation}
\setlength{\tabcolsep}{1.6pt}
\renewcommand{\arraystretch}{1.08}
\resizebox{\textwidth}{!}{
\begin{tabular}{l cc cccc cc cccc}
\toprule
\multirow{2}{*}{Variant}
& \multicolumn{2}{c}{Grounding}
& \multicolumn{4}{c}{ScanQA}
& \multicolumn{2}{c}{SQA3D}
& \multicolumn{4}{c}{Scan2Cap} \\
\cmidrule(lr){2-3}
\cmidrule(lr){4-7}
\cmidrule(lr){8-9}
\cmidrule(lr){10-13}
& ScanRefer mIoU$\uparrow$
& Multi3DRefer mIoU$\uparrow$
& C$\uparrow$
& B-4$\uparrow$
& M$\uparrow$
& R$\uparrow$
& EM$\uparrow$
& EM-R$\uparrow$
& C@.5$\uparrow$
& B-4@.5$\uparrow$
& M@.5$\uparrow$
& R@.5$\uparrow$ \\
\midrule
Full
& 47.0 & 47.9
& 95.8 & 16.8 & 19.0 & 45.0
& 55.5 & 58.3
& 79.1 & 37.1 & 27.3 & 57.8 \\
w/o $\mathcal{R}_{\mathrm{ent}}$
& 46.4 & 47.1
& 95.1 & 16.4 & 18.8 & 44.7
& 55.1 & 58.0
& 78.4 & 36.8 & 27.1 & 57.6 \\
w/o $\mathcal{R}_{\mathrm{frame}}$
& 46.2 & 47.0
& 95.3 & 16.5 & 18.8 & 44.7
& 55.0 & 58.0
& 78.6 & 36.8 & 27.1 & 57.6 \\
w/o $\mathcal{R}_{\mathrm{rel}}$
& 46.0 & 46.8
& 94.7 & 16.2 & 18.6 & 44.4
& 54.9 & 57.7
& 77.9 & 36.5 & 27.0 & 57.4 \\
\bottomrule
\end{tabular}
}
\end{table*}

Table~\ref{tab:loss_ablation} shows that removing any role-preserving objective weakens the model, while the degradation is generally smaller than directly removing the corresponding state component. 
This is expected because the state structure remains available, but its intended role is less explicitly stabilized during training. 
Removing $\mathcal{R}_{\mathrm{ent}}$ leads to mild drops on object-sensitive tasks, suggesting that entity semantic preservation helps maintain object-core evidence. 
Removing $\mathcal{R}_{\mathrm{frame}}$ affects grounding performance, indicating that frame coverage preservation helps maintain boundary and region references. 
Removing $\mathcal{R}_{\mathrm{rel}}$ causes relatively larger drops on Scan2Cap and grounding metrics, suggesting that geometry-aware relation regularization helps stabilize relation states for spatially grounded description and referring tasks.

Overall, this ablation complements the main state-component ablation. 
The main ablation studies whether each state type is needed in the representation, while this analysis studies whether the roles of existing states are better maintained by the corresponding training objectives.

\section{Limitation and Broader Impact}
\label{sec:limitation_broader_impact}

\paragraph{Limitations.}
SceneScaffold is currently evaluated mainly on indoor, static, ScanNet-style environments. Its scene-frame states rely on coarse boundary and occupied-region anchors, and its relation states capture lightweight geometry-aware interaction cues rather than complete scene graphs or full physical topology. Extending this framework to outdoor scenes, dynamic settings, long-horizon embodied interaction, and noisy real-time perception remains future work. In addition, although our ablations and functional interventions provide evidence that the states contribute complementary information, they do not fully explain how the LLM internally uses each state. More adaptive state allocation and stronger interpretability tools may further improve the framework.

\paragraph{Broader impact.}
SceneScaffold does not introduce new datasets with personal or biometric information. It may benefit embodied AI, assistive robotics, augmented reality, indoor navigation, and spatially aware human-computer interaction by improving 3D spatial grounding and reasoning. However, inaccurate 3D understanding in safety-critical or privacy-sensitive environments may lead to unsafe actions or expose sensitive indoor layouts. Practical deployment should therefore include human oversight, uncertainty estimation, fail-safe mechanisms, privacy-preserving data handling, and compliance with dataset licenses and consent requirements.

\newpage
\section*{NeurIPS Paper Checklist}

\begin{enumerate}

\item {\bf Claims}
    \item[] Question: Do the main claims made in the abstract and introduction accurately reflect the paper's contributions and scope?
    \item[] Answer: \answerYes{} 
    \item[] Justification: The abstract and introduction state the main claim that SceneScaffold reorganizes the fixed 3D visual bottleneck into role-factored scene states. The experimental section evaluates this claim on grounding, question answering, and dense captioning benchmarks, including diagnostic and ablation studies.

\item {\bf Limitations}
    \item[] Question: Does the paper discuss the limitations of the work performed by the authors?
    \item[] Answer: \answerYes{} 
    \item[] Justification: We discuss limitations in Appendix.

\item {\bf Theory assumptions and proofs}
    \item[] Question: For each theoretical result, does the paper provide the full set of assumptions and a complete (and correct) proof?
    \item[] Answer: \answerNA{} 
    \item[] Justification: The paper does not present formal theoretical results or proofs.

    \item {\bf Experimental result reproducibility}
    \item[] Question: Does the paper fully disclose all the information needed to reproduce the main experimental results of the paper to the extent that it affects the main claims and/or conclusions of the paper (regardless of whether the code and data are provided or not)?
    \item[] Answer: \answerYes{} 
    \item[] Justification: The paper specifies datasets, metrics, model backbone, token allocation, training objective, optimizer, learning rate, batch size, epochs, hardware, and evaluation protocols in the main text and appendix.

\item {\bf Open access to data and code}
    \item[] Question: Does the paper provide open access to the data and code, with sufficient instructions to faithfully reproduce the main experimental results, as described in supplemental material?
    \item[] Answer: \answerYes{} 
    \item[] Justification: We will provide these.

\item {\bf Experimental setting/details}
    \item[] Question: Does the paper specify all the training and test details (e.g., data splits, hyperparameters, how they were chosen, type of optimizer) necessary to understand the results?
    \item[] Answer: \answerYes{} 
    \item[] Justification: Section 4.1 reports the evaluated datasets, metrics, backbone, visual-token budget, state allocation, trainable modules, optimizer, learning rate, batch size, number of epochs, and hardware.

\item {\bf Experiment statistical significance}
    \item[] Question: Does the paper report error bars suitably and correctly defined or other appropriate information about the statistical significance of the experiments?
    \item[] Answer: \answerNo{} 
    \item[] Justification: We do not report error bars or confidence intervals because full multi-seed training across all 3D-LMM benchmarks is computationally expensive. We instead provide consistent results across multiple tasks, diagnostic subsets, component ablations, and functional interventions.

\item {\bf Experiments compute resources}
    \item[] Question: For each experiment, does the paper provide sufficient information on the computer resources (type of compute workers, memory, time of execution) needed to reproduce the experiments?
    \item[] Answer: \answerYes{} 
    \item[] Justification: The paper reports that experiments are trained on 8 NVIDIA A800 GPUs.
    
\item {\bf Code of ethics}
    \item[] Question: Does the research conducted in the paper conform, in every respect, with the NeurIPS Code of Ethics \url{https://neurips.cc/public/EthicsGuidelines}?
    \item[] Answer: \answerYes{} 
    \item[] Justification: The work uses standard public 3D scene understanding benchmarks and does not introduce personal, biometric, or sensitive identity data.

\item {\bf Broader impacts}
    \item[] Question: Does the paper discuss both potential positive societal impacts and negative societal impacts of the work performed?
    \item[] Answer: \answerYes{} 
    \item[] Justification: Appendix discusses potential positive applications in embodied AI, assistive robotics, as well as risks related to unsafe actions and privacy-sensitive indoor layouts.
    
\item {\bf Safeguards}
    \item[] Question: Does the paper describe safeguards that have been put in place for responsible release of data or models that have a high risk for misuse (e.g., pre-trained language models, image generators, or scraped datasets)?
    \item[] Answer: \answerNA{} 
    \item[] Justification: The paper does not release a high-risk pretrained generative model or a newly collected dataset.

\item {\bf Licenses for existing assets}
    \item[] Question: Are the creators or original owners of assets (e.g., code, data, models), used in the paper, properly credited and are the license and terms of use explicitly mentioned and properly respected?
    \item[] Answer: \answerYes{} 
    \item[] Justification: The paper cites the existing datasets, models, and benchmarks used in the experiments.

\item {\bf New assets}
    \item[] Question: Are new assets introduced in the paper well documented and is the documentation provided alongside the assets?
    \item[] Answer: \answerNA{} 
    \item[] Justification: The paper does not introduce a new dataset or benchmark.

\item {\bf Crowdsourcing and research with human subjects}
    \item[] Question: For crowdsourcing experiments and research with human subjects, does the paper include the full text of instructions given to participants and screenshots, if applicable, as well as details about compensation (if any)? 
    \item[] Answer: \answerNA{} 
    \item[] Justification: Our paper does not involve this.

\item {\bf Institutional review board (IRB) approvals or equivalent for research with human subjects}
    \item[] Question: Does the paper describe potential risks incurred by study participants, whether such risks were disclosed to the subjects, and whether Institutional Review Board (IRB) approvals (or an equivalent approval/review based on the requirements of your country or institution) were obtained?
    \item[] Answer: \answerNA{} 
    \item[] Justification: Our paper does not involve this.

\item {\bf Declaration of LLM usage}
    \item[] Question: Does the paper describe the usage of LLMs if it is an important, original, or non-standard component of the core methods in this research? Note that if the LLM is used only for writing, editing, or formatting purposes and does \emph{not} impact the core methodology, scientific rigor, or originality of the research, declaration is not required.
    \item[] Answer: \answerNA{} 
    \item[] Justification: Any use of language-model tools, if applicable, was limited to writing or formatting assistance and did not affect the core methodology, experiments, or scientific claims.

\end{enumerate}

\end{document}